\documentclass{article}
\usepackage[eabstract]{log_2025}			

\usepackage{booktabs}						
\usepackage{multirow}						
\usepackage{amsfonts}						
\usepackage{graphicx}						
\usepackage{duckuments}						
\usepackage{times}  
\usepackage{helvet}  
\usepackage{courier}  
\usepackage{wrapfig}
\usepackage{natbib}  
\usepackage{caption} 
\usepackage{algorithm}
\usepackage{algorithmic}
\usepackage{url}            
\usepackage{booktabs}       
\usepackage{amsfonts}       
\usepackage{nicefrac}       
\usepackage{microtype}      
\usepackage{xcolor}         
\usepackage{multicol}
\usepackage{multirow}
\usepackage{graphicx}
\usepackage{subfigure}
\usepackage{subcaption}
\usepackage{amsmath}
\usepackage{amssymb}

\definecolor{highlight}{RGB}{0, 0, 0}

\title[When Structure Doesn’t Help: LLMs Do Not Read Text-Attributed Graphs as Effectively as We Expected]{When Structure Doesn’t Help: LLMs Do Not Read Text-Attributed Graphs as Effectively as We Expected
}

\author[Xu et al.]{%
Haotian Xu\\
Stony Brook University \\
\email{haotian.xu@stonybrook.edu}\And
Yuning You\\
California Institute of Technology\\
\email{ynyou@caltech.edu}\And
Tengfei Ma\\
Stony Brook University\\
\email{tengfei.ma@stonybrookmedicine.edu}
}

\begin{document}

\maketitle

\begin{abstract}
Graphs provide a unified representation of semantic content and relational structure, making them a natural fit for domains such as molecular modeling, citation networks, and social graphs. Meanwhile, large language models (LLMs) have excelled at understanding natural language and integrating cross-modal signals, sparking interest in their potential for graph reasoning. Recent work has explored this by either designing template-based graph templates or using graph neural networks (GNNs) to encode structural information.In this study, we investigate how different strategies for encoding graph structure affect LLM performance on text-attributed graphs. Surprisingly, our systematic experiments reveal that: (i) \textit{LLMs leveraging only node textual descriptions already achieve strong performance across tasks}; and (ii) \textit{most structural encoding strategies offer marginal or even negative gains}. We show that explicit structural priors are often unnecessary and, in some cases, counterproductive when powerful language models are involved. This represents a significant departure from traditional graph learning paradigms and highlights the need to rethink how structure should be represented and utilized in the LLM era. \textbf{Our study is to systematically challenge the foundational assumption that structure is inherently beneficial for LLM-based graph reasoning, opening the door to new, semantics-driven approaches for graph learning.}

\end{abstract}

\section{Introduction}
\begin{figure*}[!htp]
    \centering
    \includegraphics[width=0.99\linewidth]{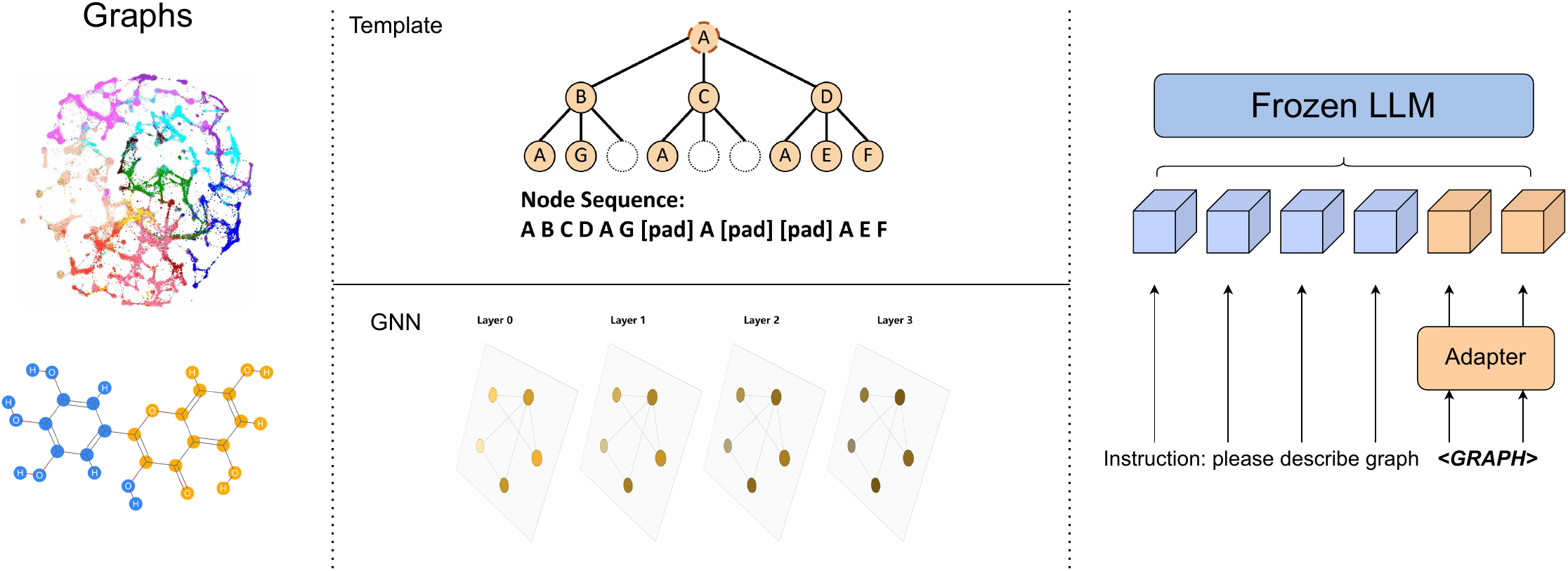}
    \caption{We present a common paradigm for aligning graph type data into LLMs. On the left, one needs to define the graph (citation network, molecule, protein, etc) and parameterize it with proper structures. In the middle, we briefly delineate the strategies encoding graphs into a LLM-favored representations: Template-based encoding will arrange each node inside graph according to a pre-defined sequence, while GNN-based encoding is to have a pretrained or random initialized GNN module to encode graphs into LLM hidden space. On the right is the pipeline to align graph modality into LLMs.}
    \label{fig:llm-graph}
\end{figure*}

Graphs are fundamental data structures for modeling relationships across diverse domains. Their capacity to capture interactions makes them invaluable for both data representation and reasoning. Over the past decade, the machine learning community has widely adopted graphs to unify multimodal data~\citep{dwivedi2022long, mccallum2000automating, Sen_Namata_Bilgic_Getoor_Galligher_Eliassi-Rad_2008}, with Graph Neural Networks (GNNs) emerging as the standard approach~\citep{kipf2017semi, velickovic2018graph, xu2018how, hamilton2017inductive, chen2018fastgcn, wang2023gmoe, muller2024attending, neubauer2024toward, ying2021do}. Recently, the rise of Large Language Models (LLMs) has opened new opportunities for integrating linguistic reasoning into graph learning, giving rise to graph foundation models.

LLM-GNN hybrids aim to combine the generalization and reasoning abilities of LLMs with the structural inductive biases of GNNs. This integration has shown promise on textual attribute graphs, where nodes carry rich semantic content. Strategies, shown in Figure~\ref{fig:llm-graph} such as prompt-based graph encoding, hybrid model architectures, and structure-aware instruction tuning have been explored~\citep{pmlr-v235-chen24bh, wang2024llms, perozzi2024letgraphtalkingencoding, he2024gretriever}. However, the role of structural information in these models remains uncertain. For example, \citet{bechler2024graph} show that GNNs may over-rely on structure even when it's irrelevant, while structure-agnostic models like DeepSets~\citep{zaheer2017deep} often generalize well. Additionally, standard graph benchmarks may fail to reflect real-world relational complexity, raising concerns about their validity~\citep{bechler2025position}.

In this work, we take a methodological perspective to re-examine the necessity of structural encodings in LLM-based graph learning. Through systematic experiments across multiple graph types, encoding templates, and modeling paradigms, we find that the inclusion of structural information, whether predefined positional encodings or message passing networks, often yields limited or no performance gains when rich semantic node features are present. In some cases, structural signals can even degrade performance due to oversmoothing or noise. We question the prevailing assumption that graph structure is inherently beneficial and suggest a shift toward more minimal, semantics-centered representations when using LLMs for graph-related tasks. \textbf{Our study calls for a rethinking of graph learning in the era of powerful language models, advocating for the design of LLM frameworks that prioritize meaningful textual context over handcrafted structural encodings.} Our code is available at \url{https://github.com/hxu105/llm-graph}.

\section{Related Work}
\textbf{Graph Learning}: Graph learning offers a flexible framework for modeling relational and structural data across domains such as social networks, biology, and knowledge graphs. At the core of this field are Graph Neural Networks (GNNs), which learn node- and graph-level representations through message passing and neighborhood aggregation~\citep{kipf2017semi, hamilton2017inductive}. Variants like Graph Attention Networks~\citep{velickovic2018graph} and spectral methods~\citep{bruna2013spectral} have been developed to address limitations in scalability and expressiveness. Inspired by advances in NLP and vision, self-supervised learning has gained popularity in the graph domain, with methods such as GraphCL~\citep{You2020GraphCL}, G-BERT~\citep{shang2019pre}, and GPT-GNN~\citep{gpt_gnn} employing contrastive or masked prediction objectives to improve generalization. However, unlike in NLP and vision, graph pretraining lacks standardized benchmarks and consistent input formats, making it harder to transfer models across domains. In response, graph foundation models (GFMs) such as GraphMAE~\citep{hou2023graphmae2}, GRAND~\citep{feng2020grand}, and GraphMVP~\citep{liu2022pretraining} aim to learn general-purpose graph representations. Despite their progress, challenges like data heterogeneity and the absence of a shared vocabulary persist—fueling growing interest in leveraging large language models (LLMs) to enhance graph representation learning.

\textbf{LLMs as GFMs}: Recent studies have advanced beyond traditional GNN-based graph foundation models (GFMs) by exploring large language models (LLMs) as graph learners, leveraging their strong generalization and multimodal capabilities. \citet{fatemi2024talk} provides a comprehensive analysis of how graph-to-text encoding strategies influence LLM performance, highlighting the importance of task type, encoding method, and graph structure. Building on this, LLaGA~\citep{pmlr-v235-chen24bh} introduces a unified framework that transforms graph data into LLM-friendly sequences using structure-aware node reordering and projection, achieving strong generalization and interpretability. PromptGFM~\citep{zhu2025llm} integrates in-text graph prompting and a learned graph vocabulary to unify GNNs and LLMs, enabling scalable and transferable reasoning on textual attribute graphs. \citet{ge-etal-2025-graph} improves graph prompting by showing that the sequential order of graph descriptions significantly affects LLM reasoning performance on graph tasks. Similarly, LLM-BP~\citep{wang2025model} enhances inference by combining task-adaptive LLM embeddings with belief propagation guided by LLM-estimated homophily scores. \textcolor{highlight}{\citet{huang2024can} investigate the role of structural information when incorporated into natural language prompts, while our work focuses on modality alignment and how LLMs internally process graph modality through adapters. Nevertheless, our findings share a similar observation with \citet{huang2024can}: LLMs tend to interpret structure-aware prompts more as contextual narratives rather than explicit topological signals. \citet{wu2025llmnodebed} introduce \textit{LLMNodeBed}, a benchmark analyzing when LLMs help in node classification across datasets and paradigms. In contrast, our work focuses on how LLMs process graph information, revealing through controlled ablations that they often act as unordered set readers when node semantics dominate, providing a mechanistic understanding rather than a benchmarking study.}

In contrast, hybrid approaches like GraphToken~\citep{perozzi2024letgraphtalkingencoding} inject structural information via GNN adapters and parameter-efficient prompts. Extensions such as G-Retriever~\citep{he2024gretriever} and TEA-GLM~\citep{wang2024llms} further integrate structural and textual features to achieve strong performance across graph-text benchmarks. SKETCH~\citep{zhou-etal-2025-taming} fuses graphs with LLMs by embedding structural and semantic aggregation into text encoding; GraphInsight~\citep{cao-etal-2025-graphinsight} mitigates positional bias through strategic placement of key graph information and RAG-style external retrieval to boost structural understanding; GALLa~\citep{zhang-etal-2025-galla} utilizes GNNs to inject code structural information as an auxiliary task. \citet{guan2025attention} investigate LLM attention patterns on graph inputs and find that transformer attention fails to align with actual graph connectivity—suggesting a gap in how LLMs internally process structural cues, rather than evaluating their downstream utility. However, most of previous works hold the assumptions that LLMs share the same inductive bias as GNNs, while we question such a belief and assess the role of structural information for LLMs processing graphs. 

\begin{table}[!htp]
    \centering
    \caption{TAG Datasets selected in experiments.}
    \begin{tabular}{ccc}
    \toprule
         Dataset & Text Domain & Graph Structure  \\
    \midrule
         Cora~\citep{mccallum2000automating} & Publication & Homophilic \\
         Citeseer~\citep{giles1998citeseer} & Publication & Homophilic \\
         Pubmed~\citep{sen2008collective} & Publication & Homophilic \\
         School~\citep{craven1998learning} & Webpage & Heterophilic \\
         Roman Empire~\citep{platonov2023critical} & Wikipedia & Heterophilic \\
         Amazon Ratings~\citep{platonov2023critical} & E-commerce & Heterophilic \\
    \bottomrule
    \end{tabular}
    \label{tab:dataset}
\end{table}
\begin{table*}[!htp]
\caption{To evaluate the utility of Laplacian embeddings for LLMs, we compare LLaGA’s ND template with our heuristic templates, HN and CO, where HN-1 samples node sequences from the 1-hop neighborhood. As shown below, explicit structural encodings do not consistently enhance performance and can even degrade it in some cases.}
\centering
\fontsize{7}{12}\selectfont
	\centering
		\begin{tabular}{cc|ccc|cc}
			\toprule
			\toprule
			\multirow{2}{*}{Setting} & \multirow{2}{*}{Dataset} & \multicolumn{3}{c}{Node Classification} & \multicolumn{2}{|c}{Link Prediction}  \cr \cmidrule(lr){3-7} & & \multirow{1}{*}{ND} & \multirow{1}{*}{HN-1} & \multirow{1}{*}{CO} & \multirow{1}{*}{ND} & \multirow{1}{*}{HN-1} \cr
                
                \midrule
                \midrule
                \multirow{3}{*}{Homophilic} & \multirow{1}{*}{Cora} & 88.07\% (0.74\%) & \textbf{88.56\%} (0.80\%) & 85.42\% (1.78\%) & 85.56\% (1.33\%) & \textbf{87.27\%} (1.56\%) \cr
                 & \multirow{1}{*}{Citeseer} & \textbf{80.31\%} (0.81\%) & 80.20\% (0.94\%) & 77.74\% (0.31\%) & 86.73\% (0.63\%) & \textbf{88.79\%} (0.84\%) \cr
                  & \multirow{1}{*}{Pubmed} & 92.56\% (0.71\%) & 94.80\% (0.17\%) & \textbf{94.84\%} (0.04\%) & 88.25\% (0.31\%) & \textbf{90.98\%} (0.38\%) \cr
                \midrule
                \multirow{3}{*}{Heterophilic} & \multirow{1}{*}{Shool} & 66.43\% (3.69\%) & 82.02\% (12.79\%) & \textbf{91.13\%} (1.66\%) & \textbf{68.61\%} (0.21\%) & 68.12\% (1.51\%) \cr
                 & \multirow{1}{*}{Roman Empire} & 48.56\% (1.17\%) & 59.70\% (2.42\%) & \textbf{62.24\%} (0.19\%) & 81.59\% (0.50\%) & \textbf{83.81\%} (0.12\%) \cr
                  & \multirow{1}{*}{Amazon Ratings} & 40.97\% (0.56\%) & \textbf{41.67\%} (0.22\%) & 40.38\% (1.14\%) & 80.26\% (2.01\%) & \textbf{84.51\%} (0.53\%) \cr
                  \midrule
                  \multicolumn{2}{c}{Across Datasets} & 69.48\% & 74.49\% & \textbf{75.29\%} & 81.83\% & \textbf{83.91\%} \cr
			\bottomrule
			\bottomrule
		\end{tabular}
\label{tab:LLaGA}
\end{table*}
\section{Do LLMs Read TAG as Expected?}

In standard graph learning, models aim to capture relationships between entities by combining semantic information, such as node features or textual descriptions, with structural information derived from graph connectivity. While node attributes provide rich local context, structural links define how entities interact within a broader topology, a dual perspective widely credited for the effectiveness of Graph Neural Networks (GNNs) across many downstream tasks. Motivated by this, recent research integrating Large Language Models (LLMs) with graphs has largely focused on injecting structural signals into LLMs. Parameter-free methods like LLaGA~\citep{pmlr-v235-chen24bh} verbalize graph structure via handcrafted templates, whereas hybrid approaches such as GraphToken, G-Retriever, and TEA-GLM~\citep{perozzi2024letgraphtalkingencoding, he2024gretriever, wang2024llms} employ GNN-based adapters to encode structure into learned embeddings, combining the relational inductive biases of GNNs with the expressive capabilities of LLMs.

These strategies generally fall into two categories: (1) template-based methods that incorporate neighbor aggregation or positional encodings, and (2) GNN-based methods that learn structural representations through neural encoders. Despite their architectural differences, both approaches often yield similar performance. In many text-rich graph tasks, the added structural information, whether hand-made or learned, contributes marginal gains or even degrades performance when strong node-level semantics are already present. This suggests that LLMs may primarily treat input graphs as unordered sets, relying more heavily on the content of selected node sequences than on the underlying graph topology. Our findings challenge the common assumption that structural information is essential for LLM-based graph modeling, and they call for a rethinking of how structure should be incorporated, if at all, into future graph foundation models for semantically rich settings.

\subsection{Preliminary}
We revisit recent LLM-Graph approaches, such as LLaGA~\citep{pmlr-v235-chen24bh} and GraphToken~\citep{perozzi2024letgraphtalkingencoding}, focusing on modality fine-tuned node classification and linke prediction in textual attribute graphs (TAGs). Our analysis is guided by two key questions: (1) Are explicit structural encodings, like Laplacian embeddings, necessary for LLMs? (2) How does message passing networks like GNNs affect performance? We conduct most of our experiments using Vicuna-7b-v1.5~\citep{zheng2023judging}.

\textbf{Datasets} As summarized in Table~\ref{tab:dataset}, we evaluate our models on six real-world TAG datasets spanning diverse text domains and structural properties. These include citation networks, e-commerce platforms, historical Wikipedia articles, and web page graphs, covering both homophilic and heterophilic patterns. Additional experiment details are provided in Appendix \ref{app:dataset}, \ref{app:exp} and \ref{app:prompts}.

\subsection{Template-Based Encoding}

In this subsection, we revisit the LLaGA framework~\citep{pmlr-v235-chen24bh}, with a particular emphasis on the \textit{Neighborhood Detail (ND)} template. This template is built upon a predefined computational graph, typically a $k$ hop B tree, and incorporates Laplacian-based positional encodings to inject structural priors into the LLM input. To rigorously evaluate the contribution of these structural components, we conduct a systematic ablation study in which both the handcrafted subgraph and the positional encodings are removed and replaced with a simple, order-invariant sequence of node descriptions.

We benchmark the original ND template against two lightweight, structure-agnostic variants: (1) \textbf{HN (Hop Neighbor)}, which randomly samples a subset of $k$-hop neighbors to construct the node sequence, and (2) \textbf{CO (Center Only)}, which provides only the description of the central node. As shown in Table~\ref{tab:LLaGA}, the ND template fails to surpass the other two structure-free templates in both node classification task and link prediction task. And including such structural embeddings can be harmful for LLMs recognizing nodes in a heterophilic graph. Surprisingly, the CO variant performs competitively, particularly on heterophilic graphs, suggesting that in some cases, including only the central node may be sufficient, and that incorporating additional neighbor context can even degrade performance.

These results indicate that for node classification on text-attributed graphs (TAGs), LLMs are often capable of extracting sufficient predictive signals from isolated node semantics, with minimal reliance on explicit structural information. This effectively transforms the graph reasoning task into a set-based problem. We observe a similar trend in link prediction tasks, where structural understanding is typically more critical. Even in this setting, augmenting the input with handcrafted structures such as Laplacian positional encodings provides limited benefit. Instead, \textbf{a simple, unordered aggregation of neighboring node descriptions enables the LLM to infer both node semantics and relational connectivity with surprising effectiveness.}

\begin{table*}[!htp]
\caption{This table evaluates whether message passing effectively aggregates useful neighbor information. Comparing a simple MLP baseline with GNN-based adapters, we find that in the LLM setting, message passing can lead to over-smoothing, even with skip connections, reducing the semantic distinctiveness of target nodes. \textbf{Best} results are bolded, \underline{second best} are underlined.}
\centering
\fontsize{8}{10}\selectfont
	\centering
		\begin{tabular}{cc|cccc}
			\toprule
			\toprule
			\multirow{2}{*}{Setting} & \multirow{2}{*}{Dataset} & \multicolumn{4}{c}{Node Classification}  \cr \cmidrule(lr){3-6} & & \multirow{1}{*}{MLP} & \multirow{1}{*}{GCN} & \multirow{1}{*}{GAT} & \multirow{1}{*}{GIN} \cr
                
                \midrule
                \midrule
                \multirow{3}{*}{Homophilic} & \multirow{1}{*}{Cora} & 87.09\% (0.66\%) & \underline{87.64}\% (0.84\%) & \textbf{88.25\%} (0.53\%) & 83.03\% (5.41\%) \cr
                 & \multirow{1}{*}{Citeseer} & 79.39\% (1.38\%) & \textbf{80.20\%} (0.13\%) & \underline{79.74\%} (0.41\%) & 79.32\% (1.11\%) \cr
                  & \multirow{1}{*}{Pubmed} & \textbf{94.76\%} (0.10\%) & \underline{92.24\%} (1.23\%) & 92.01\% (0.24\%) & 91.40\% (0.63\%) \cr
                \midrule
                \multirow{3}{*}{Heterophilic} & \multirow{1}{*}{Shool} & \textbf{90.17\%} (3.62\%) & 67.87\% (3.24\%) & 64.75\% (0.00\%) & \underline{70.02\%} (2.19\%) \cr
                 & \multirow{1}{*}{Roman Empire} & \textbf{65.39\%} (0.29\%) & 36.51\% (18.06\%) & 36.97\% (13.92\%) & \underline{46.92\%} (22.37\%) \cr
                  & \multirow{1}{*}{Amazon Ratings} & \textbf{40.78\%} (0.35\%) & 40.52\% (0.51\%) & \underline{40.71\%} (0.23\%) & 38.76\% (0.18\%) \cr
                  \midrule
                  \multicolumn{2}{c}{Across Datasets} & \textbf{76.26\%} & 67.50\% & 67.07\% & 68.24\% \cr
                  
			\bottomrule
			\bottomrule
            \cr
            \toprule
			\toprule
			\multirow{2}{*}{Setting} & \multirow{2}{*}{Dataset} & \multicolumn{4}{c}{Link Prediction}  \cr \cmidrule(lr){3-6} & & \multirow{1}{*}{MLP} & \multirow{1}{*}{GCN} & \multirow{1}{*}{GAT} & \multirow{1}{*}{GIN} \cr
                
                \midrule
                \midrule
                \multirow{3}{*}{Homophilic} & \multirow{1}{*}{Cora} & \underline{90.72\%} (0.85\%) & 90.51\% (1.19\%) & \textbf{91.05\%} (0.93\%) & 87.86\% (1.20\%) \cr
                 & \multirow{1}{*}{Citeseer} & 87.67\% (2.71\%) & \textbf{89.32\%} (0.53\%) & \underline{88.53\%} (0.46\%) & 78.34\% (1.99\%) \cr
                  & \multirow{1}{*}{Pubmed} & \textbf{89.14\%} (0.19\%) & \underline{89.11\%} (0.37\%) & 88.58\% (0.38\%) & 87.54\% (0.55\%) \cr
                \midrule
                \multirow{3}{*}{Heterophilic} & \multirow{1}{*}{Shool} & \underline{59.40\%} (1.92\%) & \underline{59.40\%} (3.26\%) & \textbf{62.78\%} (3.98\%) & 56.55\% (1.25\%) \cr
                 & \multirow{1}{*}{Roman Empire} & 51.60\% (0.62\%) & \underline{52.64\%} (0.68\%) & 51.00\% (1.02\%) & \textbf{53.63\%} (0.24\%) \cr
                  & \multirow{1}{*}{Amazon Ratings} & \textbf{72.59\%} (0.34\%) & \underline{72.10\%} (1.04\%) & 66.24\% (11.19\%) & 71.51\% (0.19\%) \cr
                  \midrule
                  \multicolumn{2}{c}{Across Datasets} & \underline{75.19\%} & \textbf{75.51\%} & 74.70\% & 72.57\% \cr
                  
			\bottomrule
			\bottomrule
		\end{tabular}
\label{tab:GraphToken}
\end{table*}

\subsection{GNN-Based Encoding}

In contrast to LLaGA’s template-based structural encoding, several recent studies~\citep{perozzi2024letgraphtalkingencoding, he2024gretriever, wang2024llms} have explored the integration of GNN-based modules to inject structural information into LLMs. To further investigate the necessity of such architectural components, we adopt the experimental setup introduced in the previous section and evaluate LLM performance in the absence of explicit structural cues. Our primary focus is on the GraphToken framework~\citep{perozzi2024letgraphtalkingencoding}, which incorporates GNNs with dynamically constructed graphs during fine-tuning, enabling a flexible and adaptive representation of structural context.

To isolate the contribution of structural modeling, we begin by evaluating the impact of different GNN backbones. Specifically, we replace the GNN with a simple multi-layer perceptron (MLP), while keeping all other components and training configurations constant. This ablation aims to determine whether semantic representations alone can sustain downstream performance without relying on graph-specific inductive biases. As reported in Table~\ref{tab:GraphToken}, although certain GNN architectures may exhibit advantages under specific domain conditions or structural regimes, the overall performance remains largely comparable. This observation aligns with findings from~\citep{perozzi2024letgraphtalkingencoding}, suggesting that the marginal gains introduced by structural modeling may not justify the added complexity.

\begin{wrapfigure}{r}{0.55\textwidth}
    \centering
    \includegraphics[width=0.99\linewidth]{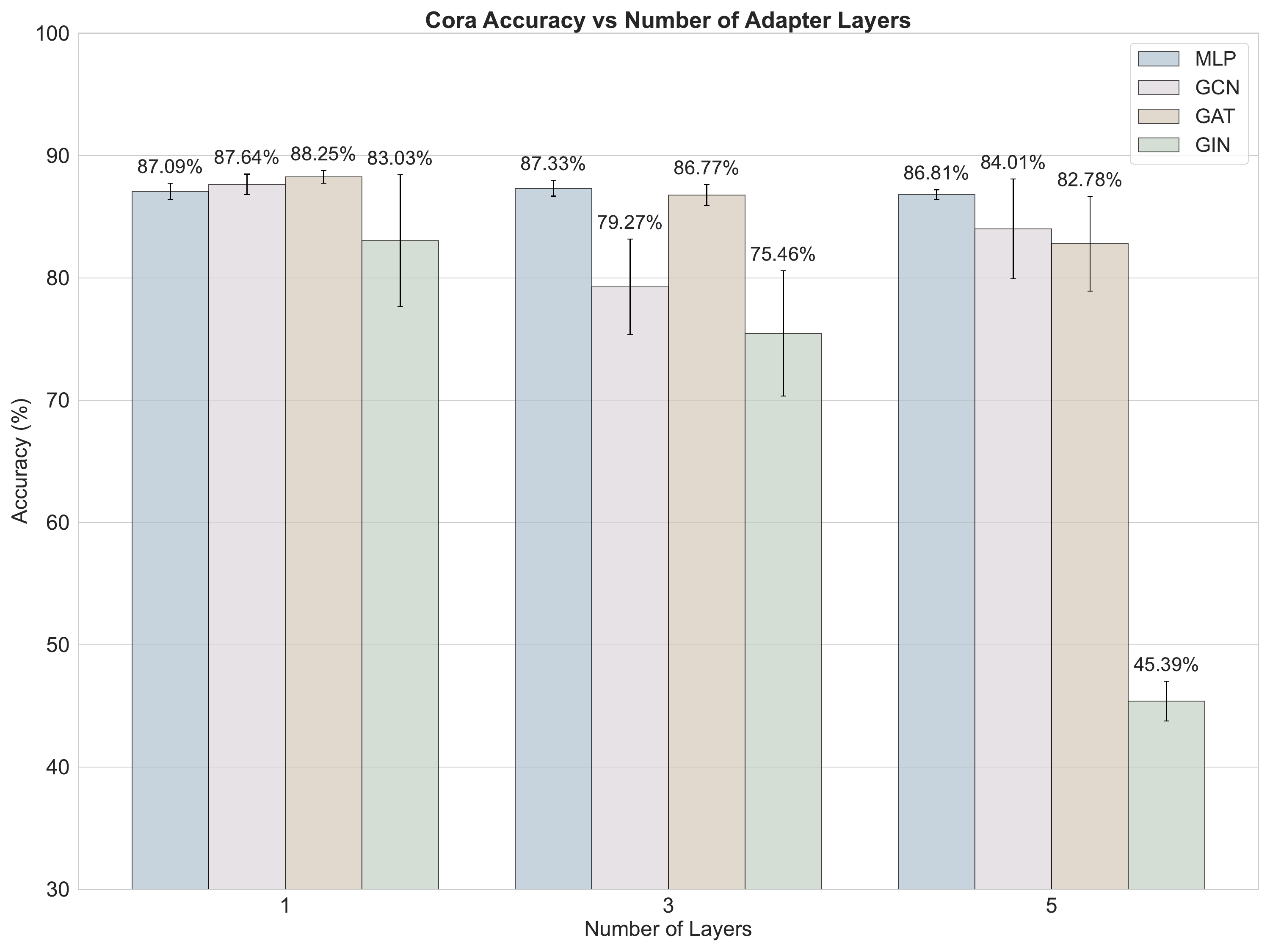}
    \caption{Increasing the number of adapter layers leads to notable performance degradation for GNN-based adapters, particularly GIN, which loses much of its generalizability in deeper configurations. In contrast, MLP adapters, without relying on structural information, maintain stable performance and exhibit greater robustness across varying depths.}
    \label{fig:depth}
\end{wrapfigure}
Furthermore, we observe that increasing the adapter depth in GraphToken consistently degrades performance when using a GNN module. As shown in Figure~\ref{fig:depth}, deeper GNN-based adapters lead to a significant fluctuation in accuracy, \textcolor{highlight}{while increasing the number of MLP layer only impacts marginally}, indicating potential overfitting or vanishing gains with deeper structural modeling.

Taken together with our earlier observations in the LLaGA setting, these results further challenge the prevailing assumption that structural encoding is critical for LLM-based graph reasoning, \textbf{suggesting that for many node classification and link prediction tasks on text-attributed graphs, LLMs can achieve strong performance by leveraging rich semantic signals alone, rendering explicit structural augmentation either redundant or even detrimental in some cases.}

\section{How Do LLMs Read Natural Graphs?}
We have previously demonstrated that structural information can be negligible or even detrimental when it interferes with node-level semantic understanding in the TAG setting. This observation aligns with the intuition that TAG connectivity is often highly correlated with the semantic descriptions of the nodes themselves. As such, LLMs may implicitly reconstruct the graph's connectivity by simply processing the node sequences. However, this raises an important question: \textit{would LLMs behave similarly on graphs that naturally exist, such as molecular structures, where topology is intrinsic rather than semantically induced?}

To investigate this, we conduct experiments on molecular property prediction, a canonical graph-level task. Specifically, we select three datasets from MoleculeNet~\citep{Ramsundar-et-al-2019}: BACE, BBBP, and HIV, chosen for their diversity in molecular properties and biomedical relevance. Full dataset statistics and preprocessing details are provided in the Appendix.

\subsection{Molecular Graphs}
Unlike TAG datasets such as citation networks or E-commerce graphs, molecular graphs are typically smaller in scale (fewer nodes) and exhibit lower average node degree, making their topological structures less complex. In such settings, template-based encoding strategies, often used to impose artificial tree-like computational paths, may introduce extraneous structural noise. Therefore, we adopt GNN-based adapters, which are more commonly used for molecular representation learning, to serve as stronger structure-aware baselines.

Interestingly, as shown in Table~\ref{tab:molecule}, even a simple MLP head applied to the embeddings of the nodes (atom), without any explicit structural modeling, can perform on par with or even outperform GNN-based adapters. This further supports our hypothesis: LLMs can extract sufficient task-relevant information from node-level semantics alone, rendering explicit structural encoding less critical for downstream performance.

To comprehensively evaluate the role of structural information in LLMs, we conducted experiments across three representative tasks. Across all three, the results consistently suggest that LLMs can operate effectively without leveraging explicit structural information, provided that high-quality node embeddings are available. Notably, the node representations used in our experiments are derived from a pretrained language encoder, ensuring rich semantic content.
\subsection{Pretrained Graph Encoder v.s. Pretrained Language Encoder}
An intriguing follow-up question emerges: what if we replace the language encoder with a pretrained graph encoder? Will structural information, as captured by the graph encoder, play a more central role in enhancing LLM performance?

\begin{wrapfigure}{l}{0.49\textwidth}
    \centering
    \caption{How Pretrained Encoders Impact}
    \includegraphics[width=0.99\linewidth]{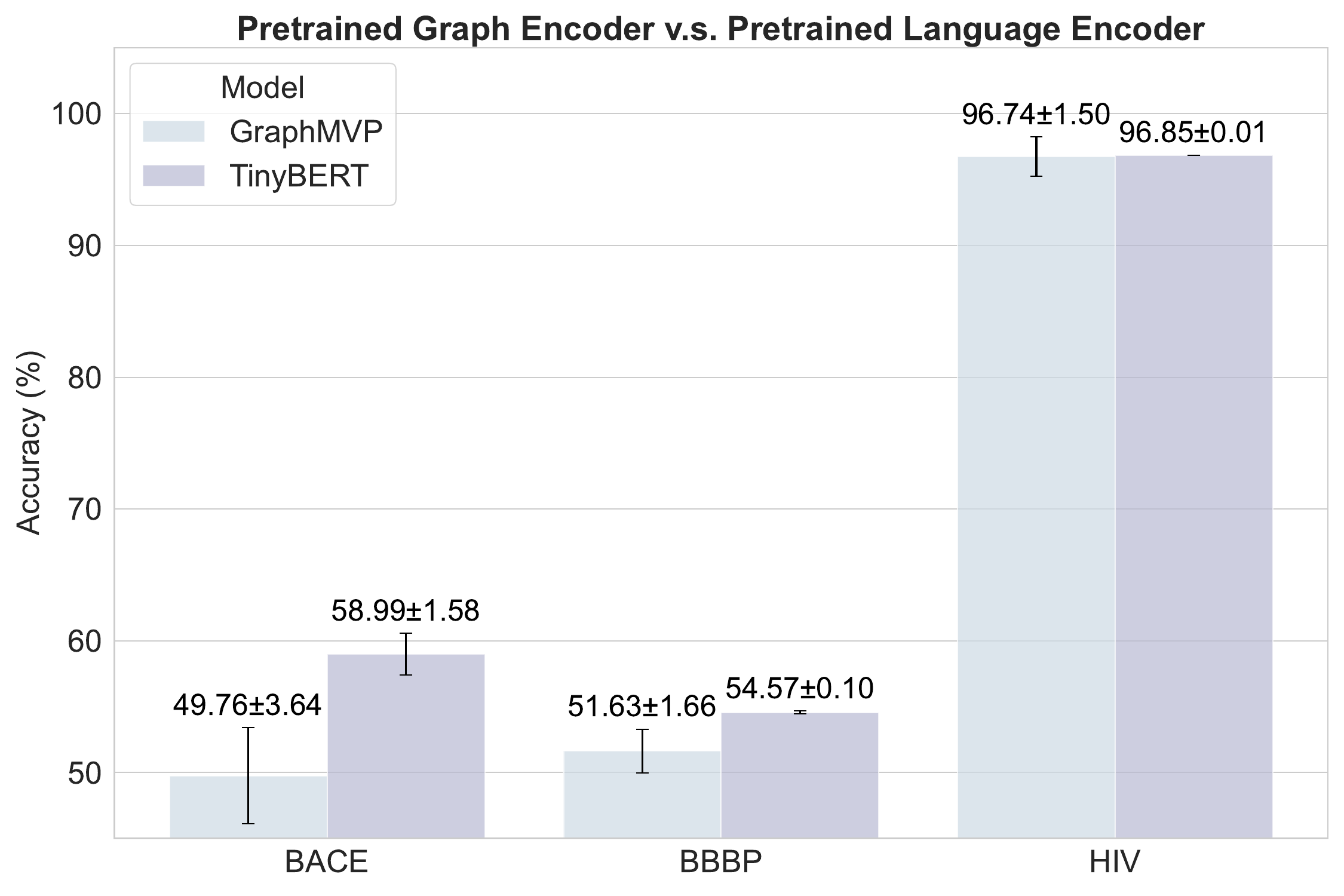}
    \label{fig:pretrained}
\end{wrapfigure}

To further investigate the role of pretrained modality-specific encoders in processing naturally occurring graphs such as molecular structures, we compare embeddings from GraphMVP~\citep{liu2022pretraining}, a state-of-the-art graph pretraining framework for molecules, against those from TinyBERT~\citep{jiao2019tinybert}, a compact yet effective pretrained language model. For a fair comparison in representation capacity, we match the embedding dimensionalities, using a 5-layer, 300-dimensional GraphMVP and a 4-layer, 312-dimensional TinyBERT. Figure~\ref{fig:pretrained} reports the average accuracy along with standard deviations across multiple runs. The results indicate that even in domains where structural priors are intrinsic, such as chemistry, pretrained graph encoders like GraphMVP do not consistently demonstrate a clear advantage in leveraging structural information for LLM-based processing. In contrast, a lightweight pretrained language encoder such as TinyBERT is sufficient to represent molecular graphs solely from sequences of atom-level descriptions, reinforcing our earlier conclusion that LLMs predominantly exploit semantic content rather than explicit structural cues.

This experiment further reinforces our central finding: LLMs tend to prioritize semantic content over structural information when processing graph-related inputs. Even when structural signals are provided through specialized graph encoders, they fail to surpass the semantic richness embedded within language-based representations. Consequently, our observations point to a broader implication: \textbf{the quality and expressiveness of semantic embeddings, rather than explicit graph topology, serve as the dominant factors determining LLM performance on graph-centric reasoning tasks}. This also challenges the conventional assumption that graph-specific pretraining inherently offers a representational advantage in capturing relational and compositional patterns.

\begin{table*}[!t]
\caption{We further investigate whether structural information provides tangible benefits for LLMs in processing graph-structured data by evaluating three molecular property prediction datasets. Consistent with our earlier findings on TAGs, we observe that plain node embeddings, devoid of any explicit structural encoding, can achieve comparable or even superior performance to structure-aware approaches on molecular tasks.}
\centering
\fontsize{9}{10}\selectfont
	\centering
		\begin{tabular}{c|cccc}
			\toprule
			\toprule
			\multirow{2}{*}{Dataset} & \multicolumn{4}{c}{Molecular Property Prediction}  \cr \cmidrule(lr){2-5} & \multirow{1}{*}{MLP} & \multirow{1}{*}{GCN} & \multirow{1}{*}{GIN} & \multirow{1}{*}{GAT} \cr
                
                \midrule
                \midrule
                \multirow{1}{*}{BACE} & \textbf{58.99\%} (1.66\%) & \underline{58.77\%} (9.13\%) & \textbf{58.99\%} (5.52\%) & 57.46\% (3.62\%) \cr
                 \multirow{1}{*}{BBBP} & 54.57\% (1.38\%) & \underline{57.84\%} (0.49\%) & \textbf{60.29\%} (0.49\%) & 51.96\% (1.47\%) \cr
                  \multirow{1}{*}{HIV} & \textbf{96.85\%} (0.01\%) & 96.81\% (0.03\%) & 96.79\% (0.00\%) & \underline{96.82\%} (0.03\%) \cr
			\bottomrule
			\bottomrule
		\end{tabular}
\label{tab:molecule}
\end{table*}

\subsection{\textcolor{highlight}{LLMs in Geometric Deep Learning}}

The current graph benchmarks may not fully capture tasks requiring genuine relational reasoning. In fact, our findings are consistent with \citet{bechler2025position}, which argues that existing graph benchmarks are limited and often fail to reflect real-world relational complexity. Our results reinforce this position by empirically demonstrating that structural information contributes marginally even in these canonical benchmarks—suggesting that many current datasets may not require, or even reward, structural reasoning.
\begin{wrapfigure}{r}{0.51\textwidth}
    \centering
    \caption{Features for LLMs on GDL.}
    \includegraphics[width=0.99\linewidth]{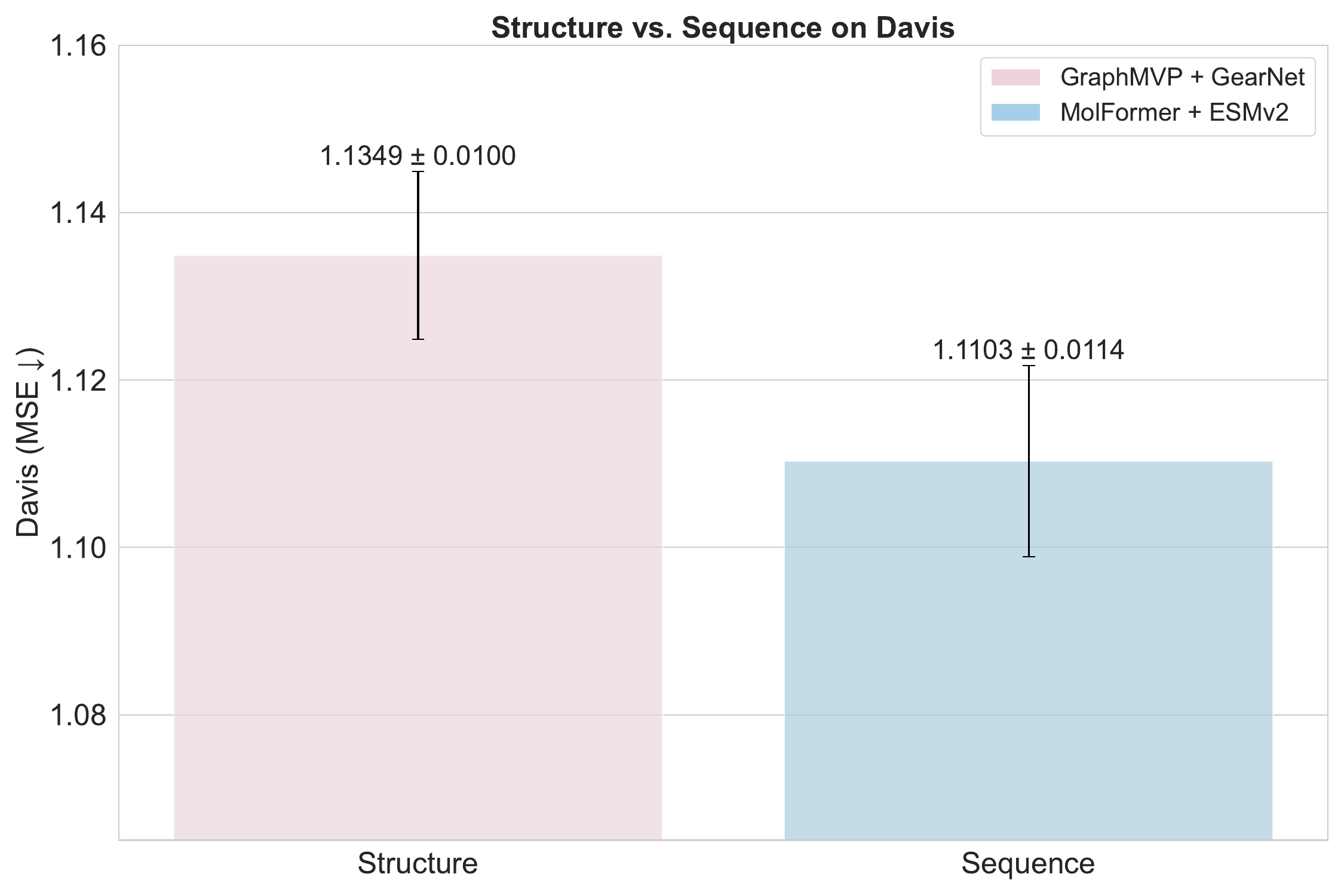}
    \label{fig:gdl}
\end{wrapfigure}
To further test our hypothesis under more structurally demanding conditions, we extended our experiments to the Davis Drug–Target Interaction (DTI) dataset, a standard benchmark in geometric deep learning (GDL) that explicitly involves molecular–protein structural reasoning. We compared performance between structure-based and sequence-based encoders, we use Llama3-8B as the LLM backbone in this experiment. DTI dataset contains molecules and proteins, and we choose GraphMVP \citep{liu2022pretraining} and GearNet \citep{zhang2022protein} to encode the molecule and protein structure representations, while for sequence information, we select MolFormer \citep{10.1038/s42256-022-00580-7} and ESMv2 \citep{lin2022language} as encoders respectively.

 As shown in Figure \ref{fig:gdl}, the sequence-based (semantic) representations perform comparably or even slightly better than structural encodings, reinforcing that our observation generalizes to more challenging, structure-dependent domains. We emphasize that our work and \citet{bechler2025position} share a similar perspective: the field should move toward rethinking graph benchmarks and embracing semantics- or geometry-aware tasks as complementary directions for future graph learning research.

\section{What affects LLMs in understanding Graphs?}

While our experiments suggest that LLMs may not inherently benefit from explicit structural information, it is important to recognize that their ability to leverage such signals can vary significantly depending on factors such as pretraining corpus, optimization strategy, and model scale. To assess the robustness of our findings, we further investigate whether the observed trends hold consistently across different backbone LLM architectures and parameter sizes. This analysis aims to disentangle model-specific artifacts from generalizable behavior, and to evaluate whether the limited utility of structural encodings persists regardless of underlying model configurations.

\subsection{Scaling Ineffectiveness}

It is widely acknowledged that increasing the parameter size of LLMs often leads to enhanced expressive power and improved performance across a broad range of tasks. To assess whether this scaling trend extends to graph-related tasks, we evaluate the impact of model size on the ability of LLMs to utilize structural information. Specifically, in Table~\ref{tab:scaling}, we compare the structure-aware ND template with the structure-free HN template using two model variants: LLaMA2-7B and LLaMA2-13B~\citep{touvron2023llama}.

Our results reinforce the patterns observed in previous experiments. Despite increasing the backbone model size, the tendency of LLMs to overlook explicit structural encodings remains consistent. Notably, scaling up to 13B parameters does not enhance the model’s ability to leverage structural information. In fact, in most cases, the structure-free HN template outperforms the structure-aware ND template, further suggesting that model scale alone does not improve sensitivity to structural signals in graph-based tasks.

\begin{table}[!htp]
\caption{Switching LLM backbones preserves our finding that structure may be unnecessary for LLMs processing graphs. Even with weak semantic content, LLMs still reveal the same pattern.}
\centering
\fontsize{8.8}{12}\selectfont
	\centering
		\begin{tabular}{cc|cc|cc}
			\toprule
			\toprule
			\multirow{2}{*}{Model Architecture} & \multirow{2}{*}{Dataset} & \multicolumn{2}{c}{Node Classification}  & \multicolumn{2}{c}{Link Prediction} \cr \cmidrule(lr){3-6} & & \multirow{1}{*}{ND} & \multirow{1}{*}{HN-1} & \multirow{1}{*}{ND} & \multirow{1}{*}{HN-1}\cr
                
                \midrule
                \midrule
                \multirow{2}{*}{Llama2-7B} & \multirow{1}{*}{Cora} & 87.76\%(0.21\%) & \textbf{88.01\%}(0.56\%) & 85.48\%(0.38\%) & \textbf{87.04\%}(0.75\%) \cr
                 & \multirow{1}{*}{School} & 70.98\%(0.83\%) & \textbf{92.09\%}(2.49\%) & 61.82\%(2.88\%) & \textbf{69.09\%}(1.92\%) \cr
                \midrule
                 \multirow{2}{*}{Llama2-13B} & \multirow{1}{*}{Cora} & \textbf{87.58\%}(0.59\%) & 87.45\%(0.19\%) & 84.24\%(0.89\%) & \textbf{86.05\%}(0.55\%) \cr
                 & \multirow{1}{*}{School} & 69.30\%(3.24\%) & \textbf{89.45\%}(3.40\%) & 61.21\%(1.28\%) & \textbf{67.15\%}(1.52\%) \cr
			\bottomrule
			\bottomrule
            \cr
            \toprule
			\toprule
			\multirow{2}{*}{Semantic Content} & \multirow{2}{*}{Dataset} & \multicolumn{2}{c}{Node Classification} & \multicolumn{2}{c}{Link Prediction} \cr \cmidrule(lr){3-6} & & \multirow{1}{*}{ND} & \multirow{1}{*}{HN-1}& \multirow{1}{*}{ND} & \multirow{1}{*}{HN-1}\cr
                
                \midrule
                \midrule
                \multirow{2}{*}{sparse} & \multirow{1}{*}{Cora} & \textbf{83.96\%}(2.74\%) & 82.17\%(0.56\%) & 69.19\%(1.15\%) & \textbf{74.81\%}(0.85\%)  \cr
                 & \multirow{1}{*}{School} & 56.95\%(6.19\%) & \textbf{73.62\%}(7.21\%) & 63.63\%(0.63\%) & \textbf{65.09\%}(3.93\%)  \cr
                \midrule
                 \multirow{2}{*}{full} & \multirow{1}{*}{Cora} & 83.39\%(0.37\%) & \textbf{84.81\%}(0.46\%) & 70.81\%(1.89\%) & \textbf{75.84\%}(0.74\%) \cr
                 & \multirow{1}{*}{School} & 59.47\%(3.97\%) & \textbf{60.19\%}(1.10\%) & 63.15\%(5.91\%) & \textbf{70.06\%}(3.30\%) \cr
			\bottomrule
			\bottomrule
		\end{tabular}
\label{tab:scaling}
\label{tab:semantic}
\end{table}

\subsection{Semantic Content}

To further assess the robustness of our findings, we investigate whether the reliance on structural information changes under weaker semantic content. Specifically, we reduce the descriptive richness of each node by comparing two settings: (1) full node descriptions, such as full abstracts or complete webpage content, and (2) sparse descriptions, limited to titles of papers or webpages. We generate node embeddings using three widely used pretrained models: RoBERTa-large~\citep{DBLP:journals/corr/abs-1907-11692}, BERT-large~\citep{DBLP:journals/corr/abs-1810-04805}, and T5-XXL~\citep{2020t5}.

Even under reduced semantic conditions, the structure-free HN template consistently matches or outperforms the structure-aware ND template (see Table~\ref{tab:semantic}). These results suggest that LLMs are capable of extracting meaningful relational patterns from minimal semantic cues, without the need for explicit structural encodings. This further reinforces our conclusion that structural augmentation provides limited benefits, even when node-level semantics are sparse.

\subsection{\textcolor{highlight}{Will Dataset Size Impact This Finding?}}
\begin{table}[!htp]
\caption{Our findings hold consistently on larger text-attributed graphs, suggesting that structural information contributes only marginally to LLMs’ graph inference.}
\centering
\fontsize{10}{12}\selectfont
	\centering
		\begin{tabular}{c|ccc}
			\toprule
			\toprule
			\multirow{2}{*}{Dataset} & \multicolumn{3}{c}{Node Classification} \cr \cmidrule(lr){2-4} &  \multirow{1}{*}{ND} & \multirow{1}{*}{HN-1}& \multirow{1}{*}{CO}\cr
            \midrule
                
                Products & 83.45\% (0.39\%) & \textbf{83.87\%} (0.24\%)	 & 80.10\% (0.27\%)\cr
                ArXiv & \textbf{75.65\%} (0.50\%) & 75.41\% (0.21\%) & 74.46\% (0.18\%)\cr
                
			\bottomrule
			\bottomrule
		\end{tabular}
\label{tab:larger-dataset}
\end{table}
To assess whether structural information benefits LLMs on larger graph datasets, we compared node classification performance under two settings: ND (structure-aware) and HN-1 (structure-agnostic). Experiments were conducted on the \textsc{Products}~\citep{Bhatia16} and \textsc{ArXiv}~\citep{wang2020microsoft} benchmarks (Table~\ref{tab:larger-dataset}), with the null hypothesis that both settings yield comparable mean performance. Each configuration was evaluated over three runs. The results show no statistically significant difference between ND and HN-1 on either dataset (two-sample Welch’s t-test: $p_{\text{products}} = 0.20 > 0.05$, $p_{\text{arxiv}} = 0.49 > 0.05$), indicating that structural information does not produce consistent gains. This finding aligns with our earlier results, suggesting that LLMs effectively capture relational dependencies without relying on explicit graph topology.

\subsection{How about Large Reasoning Models?}

Large Reasoning Models (LRMs) aim to replicate human-like problem solving by drawing conclusions from structured rules and evidence. In this context, template-based graph encodings can be viewed as implicit reasoning prompts that guide the model's attention. This motivates us to explore whether explicit structural signals, like Laplacian positional encodings, can enhance reasoning performance. We conduct a preliminary evaluation using OpenReasoning-Nemotron-7B~\citep{ahmad2025opencodereasoning}, a large-scale model post-trained for reasoning in math, science, and code domains on Cora and School datasets.
\begin{figure}[!h]
    \centering
    \caption{\textit{Left}: Though reasoning model can perform structured decision-making, it does not rely on structure information. \textit{Right}: Altering the node sequence via GDC can gain some enhancement at a time.}
    \begin{subfigure}
    \centering
    \includegraphics[width=0.49\linewidth]{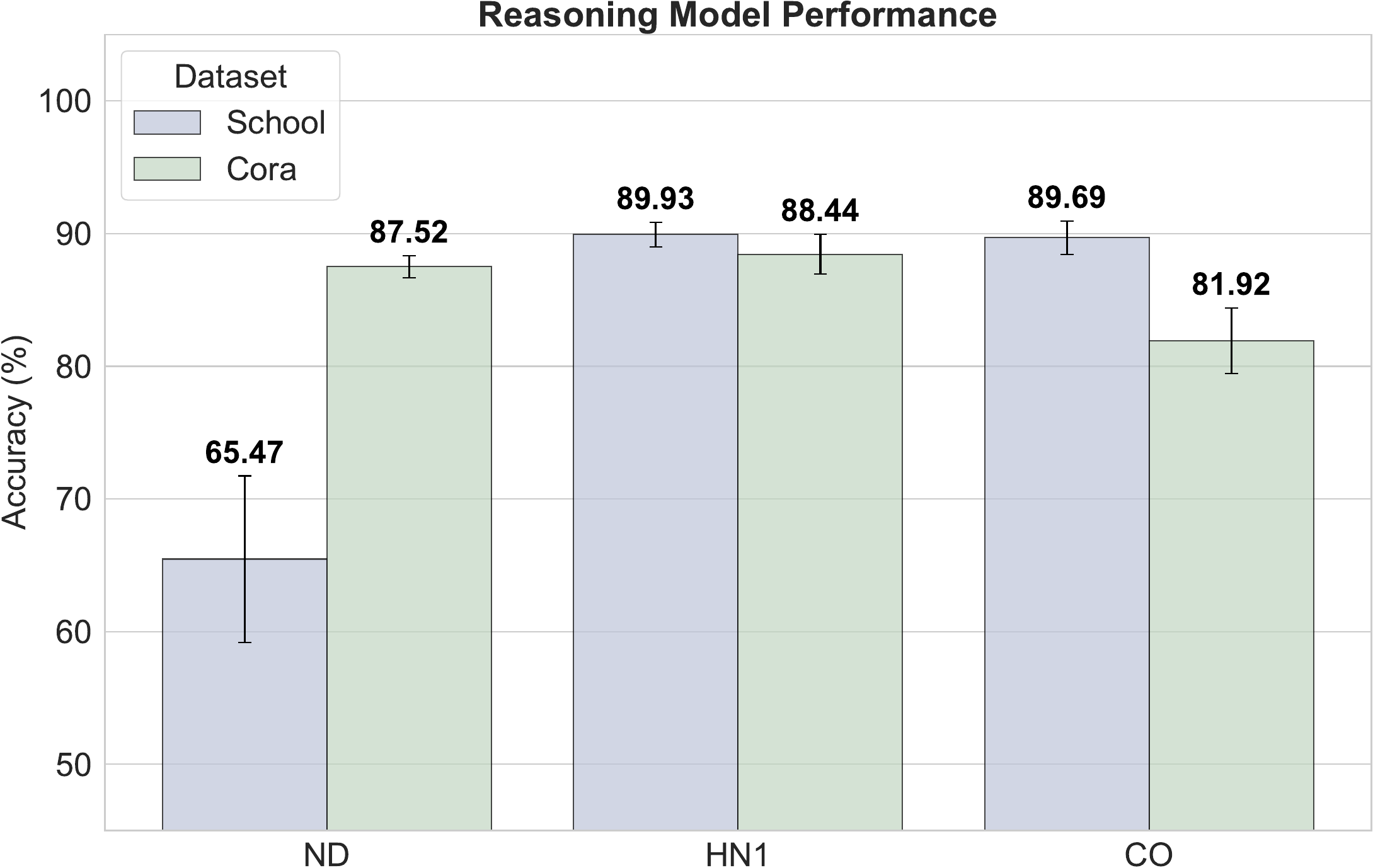}
    \label{fig:reasoning}
\end{subfigure}
\begin{subfigure}
    \centering
    \includegraphics[width=0.49\linewidth]{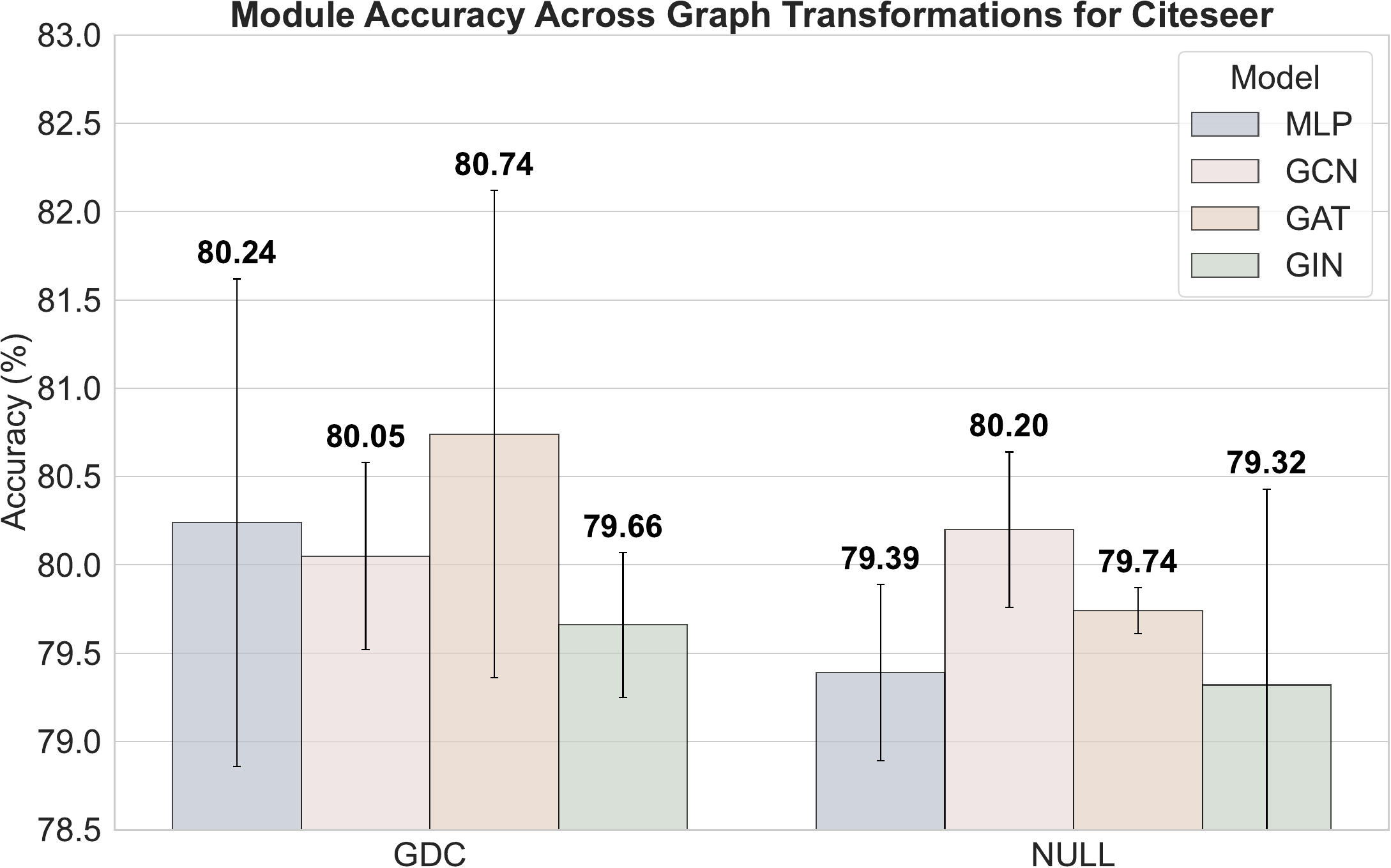}
    \label{fig:gdc}
\end{subfigure}
\end{figure}

Consistent with our earlier observations, we find that LRMs, despite their specialized training for structured reasoning, \textbf{do not exhibit significant improvements when structural encodings such as Laplacian embeddings are included}. These results further reinforce our central finding: even in models explicitly designed for reasoning, the addition of graph structural signals does not necessarily translate to better generalization or task performance. This suggests that the current generation of reasoning models primarily rely on semantic representations and may underutilize explicit graph structure unless such reasoning is explicitly aligned with the model’s pretraining or task formulation.
\section{Can LLMs Better Leverage Structure?}

Although structural information appears to have limited influence on LLM performance, we observed that the GraphToken framework, when paired with an MLP adapter, occasionally outperforms the structure-free CO template. While MLPs lack explicit message-passing mechanisms, they may still benefit from implicit structural cues preserved in the ordering of the input node sequence. Inspired by this observation, we hypothesize that optimizing node sequence selection can better expose latent structural signals to the LLM and potentially improve the performance.

To investigate this, we did a preliminary study to incorporate Graph Diffusion Convolution (GDC)~\citep{gasteiger2019diffusion}, a graph transformation technique designed to capture long-range dependencies using a sparsified generalized form of graph diffusion. GCD implicitly generates a new graph by graph diffusion as well as a following sparsification step, so the information can be aggregated from a larger neighborhood. As a result, applying GDC can also effectively condenses the input node sequence into a sparse, center-focused subset, resulting in improved LLM performance in most cases (as illustrated in Figure~\ref{fig:gdc}). While this does not overturn our main finding, it highlights a promising direction: certain graph transformations, especially those capturing longer-range information, may provide a structured yet minimal signal that LLMs can exploit more effectively. Future work may further investigate how to integrate such transformations with semantic guidance to better align graph structure with LLM capabilities.

\section{Conclusion and Future Directions}
In this study, we revisited LLM-based approaches to TAG tasks and systematically evaluated different structural encoding strategies. We find that LLMs largely treat graphs as unordered sets, showing minimal sensitivity to explicit structural cues from input templates or model-level components such as GNNs. These findings challenge the conventional view that structural information is essential for graph reasoning, highlighting instead the dominant role of semantics in LLM-based graph learning. \textbf{Our results provide an empirical foundation for understanding LLM–graph interactions and underscore the importance of effective node sequencing over structural encodings in advancing LLM performance on TAG tasks.}

\bibliographystyle{unsrtnat}
\bibliography{reference}

@InProceedings{pmlr-v235-chen24bh,
  title = 	 {{LL}a{GA}: Large Language and Graph Assistant},
  author =       {Chen, Runjin and Zhao, Tong and Jaiswal, Ajay Kumar and Shah, Neil and Wang, Zhangyang},
  booktitle = 	 {Proceedings of the 41st International Conference on Machine Learning},
  pages = 	 {7809--7823},
  year = 	 {2024},
  editor = 	 {Salakhutdinov, Ruslan and Kolter, Zico and Heller, Katherine and Weller, Adrian and Oliver, Nuria and Scarlett, Jonathan and Berkenkamp, Felix},
  volume = 	 {235},
  series = 	 {Proceedings of Machine Learning Research},
  month = 	 {21--27 Jul},
  publisher =    {PMLR},
  url = 	 {https://proceedings.mlr.press/v235/chen24bh.html},
  }

@inproceedings{
wang2024llms,
title={{LLM}s as Zero-shot Graph Learners: Alignment of {GNN} Representations with {LLM} Token Embeddings},
author={Duo Wang and Yuan Zuo and Fengzhi Li and Junjie Wu},
booktitle={The Thirty-eighth Annual Conference on Neural Information Processing Systems},
year={2024},
url={https://openreview.net/forum?id=32g9BWTndc}
}

@misc{perozzi2024letgraphtalkingencoding,
      title={Let Your Graph Do the Talking: Encoding Structured Data for LLMs}, 
      author={Bryan Perozzi and Bahare Fatemi and Dustin Zelle and Anton Tsitsulin and Mehran Kazemi and Rami Al-Rfou and Jonathan Halcrow},
      year={2024},
      eprint={2402.05862},
      archivePrefix={arXiv},
      primaryClass={cs.LG},
      url={https://arxiv.org/abs/2402.05862}, 
}

@inproceedings{
he2024gretriever,
title={G-Retriever: Retrieval-Augmented Generation for Textual Graph Understanding and Question Answering},
author={Xiaoxin He and Yijun Tian and Yifei Sun and Nitesh V Chawla and Thomas Laurent and Yann LeCun and Xavier Bresson and Bryan Hooi},
booktitle={The Thirty-eighth Annual Conference on Neural Information Processing Systems},
year={2024},
url={https://openreview.net/forum?id=MPJ3oXtTZl}
}

@inproceedings{bechler2024graph,
  title={Graph neural networks use graphs when they shouldn't},
  author={Bechler-Speicher, Maya and Amos, Ido and Gilad-Bachrach, Ran and Globerson, Amir},
  booktitle={Forty-first International Conference on Machine Learning},
  year={2024}
}

@article{bechler2025position,
  title={Position: Graph Learning Will Lose Relevance Due To Poor Benchmarks},
  author={Bechler-Speicher, Maya and Finkelshtein, Ben and Frasca, Fabrizio and M{\"u}ller, Luis and T{\"o}nshoff, Jan and Siraudin, Antoine and Zaverkin, Viktor and Bronstein, Michael M and Niepert, Mathias and Perozzi, Bryan and others},
  journal={arXiv preprint arXiv:2502.14546},
  year={2025}
}

@article{dwivedi2022long,
  title={Long range graph benchmark},
  author={Dwivedi, Vijay Prakash and Ramp{\'a}{\v{s}}ek, Ladislav and Galkin, Michael and Parviz, Ali and Wolf, Guy and Luu, Anh Tuan and Beaini, Dominique},
  journal={Advances in Neural Information Processing Systems},
  volume={35},
  pages={22326--22340},
  year={2022}
}

@article{mccallum2000automating,
  title={Automating the construction of internet portals with machine learning},
  author={McCallum, Andrew Kachites and Nigam, Kamal and Rennie, Jason and Seymore, Kristie},
  journal={Information Retrieval},
  volume={3},
  pages={127--163},
  year={2000},
  publisher={Springer}
}

@article{Sen_Namata_Bilgic_Getoor_Galligher_Eliassi-Rad_2008, title={Collective Classification in Network Data}, volume={29}, url={https://ojs.aaai.org/aimagazine/index.php/aimagazine/article/view/2157}, DOI={10.1609/aimag.v29i3.2157}, abstractNote={Many real-world applications produce networked data such as the world-wide web (hypertext documents connected via hyperlinks), social networks (for example, people connected by friendship links), communication networks (computers connected via communication links) and biological networks (for example, protein interaction networks). A recent focus in machine learning research has been to extend traditional machine learning classification techniques to classify nodes in such networks. In this article, we provide a brief introduction to this area of research and how it has progressed during the past decade. We introduce four of the most widely used inference algorithms for classifying networked data and empirically compare them on both synthetic and real-world data.}, number={3}, journal={AI Magazine}, author={Sen, Prithviraj and Namata, Galileo and Bilgic, Mustafa and Getoor, Lise and Galligher, Brian and Eliassi-Rad, Tina}, year={2008}, month={Sep.}, pages={93} }

@inproceedings{kipf2017semi,
  title={Semi-Supervised Classification with Graph Convolutional Networks},
  author={Kipf, Thomas N. and Welling, Max},
  booktitle={International Conference on Learning Representations (ICLR)},
  year={2017}
}

@article{
  velickovic2018graph,
  title="{Graph Attention Networks}",
  author={Veli{\v{c}}kovi{\'{c}}, Petar and Cucurull, Guillem and Casanova, Arantxa and Romero, Adriana and Li{\`{o}}, Pietro and Bengio, Yoshua},
  journal={International Conference on Learning Representations},
  year={2018},
  url={https://openreview.net/forum?id=rJXMpikCZ},
  note={accepted as poster},
}

@inproceedings{
xu2018how,
title={How Powerful are Graph Neural Networks?},
author={Keyulu Xu and Weihua Hu and Jure Leskovec and Stefanie Jegelka},
booktitle={International Conference on Learning Representations},
year={2019},
url={https://openreview.net/forum?id=ryGs6iA5Km},
}

@inproceedings{
chen2018fastgcn,
title={Fast{GCN}: Fast Learning with Graph Convolutional Networks via Importance Sampling},
author={Jie Chen and Tengfei Ma and Cao Xiao},
booktitle={International Conference on Learning Representations},
year={2018},
url={https://openreview.net/forum?id=rytstxWAW},
}

@inproceedings{wang2023gmoe,
author = {Wang, Haotao and Jiang, Ziyu and You, Yuning and Han, Yan and Liu, Gaowen and Srinivasa, Jayanth and Kompella, Ramana Rao  and Wang, Zhangyang},
title = {Graph Mixture of Experts: Learning on Large-Scale Graphs with Explicit Diversity Modeling},
booktitle = {NeurIPS}, 
year = {2023}
}

@article{
muller2024attending,
title={Attending to Graph Transformers},
author={Luis M{\"u}ller and Mikhail Galkin and Christopher Morris and Ladislav Ramp{\'a}{\v{s}}ek},
journal={Transactions on Machine Learning Research},
issn={2835-8856},
year={2024},
url={https://openreview.net/forum?id=HhbqHBBrfZ},
note={}
}

@misc{
neubauer2024toward,
title={Toward Principled Transformers for Knowledge Tracing},
author={Kai Neubauer and Yannick Rudolph and Ulf Brefeld},
year={2024},
url={https://openreview.net/forum?id=4dtwyV7XyW}
}

@inproceedings{
ying2021do,
title={Do Transformers Really Perform Badly for Graph Representation?},
author={Chengxuan Ying and Tianle Cai and Shengjie Luo and Shuxin Zheng and Guolin Ke and Di He and Yanming Shen and Tie-Yan Liu},
booktitle={Thirty-Fifth Conference on Neural Information Processing Systems},
year={2021},
url={https://openreview.net/forum?id=OeWooOxFwDa}
}

@article{zaheer2017deep,
  title={Deep sets},
  author={Zaheer, Manzil and Kottur, Satwik and Ravanbakhsh, Siamak and Poczos, Barnabas and Salakhutdinov, Russ R and Smola, Alexander J},
  journal={Advances in neural information processing systems},
  volume={30},
  year={2017}
}

@article{hamilton2017inductive,
  title={Inductive representation learning on large graphs},
  author={Hamilton, Will and Ying, Zhitao and Leskovec, Jure},
  journal={Advances in neural information processing systems},
  volume={30},
  year={2017}
}

@article{bruna2013spectral,
  title={Spectral networks and locally connected networks on graphs},
  author={Bruna, Joan and Zaremba, Wojciech and Szlam, Arthur and LeCun, Yann},
  journal={arXiv preprint arXiv:1312.6203},
  year={2013}
}

@inproceedings{You2020GraphCL,
 author = {You, Yuning and Chen, Tianlong and Sui, Yongduo and Chen, Ting and Wang, Zhangyang and Shen, Yang},
 booktitle = {Advances in Neural Information Processing Systems},
 editor = {H. Larochelle and M. Ranzato and R. Hadsell and M. F. Balcan and H. Lin},
 pages = {5812--5823},
 publisher = {Curran Associates, Inc.},
 title = {Graph Contrastive Learning with Augmentations},
 url = {https://proceedings.neurips.cc/paper/2020/file/3fe230348e9a12c13120749e3f9fa4cd-Paper.pdf},
 volume = {33},
 year = {2020}
}

@article{shang2019pre,
  title={Pre-training of Graph Augmented Transformers for Medication Recommendation},
  author={Shang, Junyuan and Ma, Tengfei and Xiao, Cao and Sun, Jimeng},
  journal={arXiv preprint arXiv:1906.00346},
  year={2019}
}

@inproceedings{gpt_gnn,
  title={GPT-GNN: Generative Pre-Training of Graph Neural Networks},
  author={Ziniu Hu and Yuxiao Dong and Kuansan Wang and Kai-Wei Chang and Yizhou Sun},
  booktitle={Proceedings of the 26th ACM SIGKDD Conference on Knowledge Discovery and Data Mining},
  year={2020}
}

@inproceedings{hou2023graphmae2,
  title={GraphMAE2: A Decoding-Enhanced Masked Self-Supervised Graph Learner},
  author={ Hou, Zhenyu and  He, Yufei and  Cen, Yukuo and Liu, Xiao and Dong, Yuxiao and Kharlamov, Evgeny and Tang, Jie},
  booktitle={Proceedings of the ACM Web Conference 2023 (WWW’23)},
  year={2023}
}

@inproceedings{feng2020grand,
  title={Graph Random Neural Network for Semi-Supervised Learning on Graphs},
  author={Feng, Wenzheng and Zhang, Jie and Dong, Yuxiao and Han, Yu and Luan, Huanbo and Xu, Qian and Yang, Qiang and Kharlamov, Evgeny and Tang, Jie},
  booktitle={NeurIPS'20},
  year={2020}
}

@inproceedings{liu2022pretraining,
    title={Pre-training Molecular Graph Representation with 3D Geometry},
    author={Shengchao Liu and Hanchen Wang and Weiyang Liu and Joan Lasenby and Hongyu Guo and Jian Tang},
    booktitle={International Conference on Learning Representations},
    year={2022},
    url={https://openreview.net/forum?id=xQUe1pOKPam}
}

@inproceedigs{fatemi2024talk,
  title={Talk like a Graph: Encoding Graphs for Large Language Models},
  author={Bahare Fatemi and Jonathan Halcrow and Bryan Perozzi},
  booktitle={International Conference on Learning Representations (ICLR)},
  year={2024}
}

@article{zhu2025llm,
  title={LLM as GNN: Graph Vocabulary Learning for Text-Attributed Graph Foundation Models},
  author={Zhu, Xi and Xue, Haochen and Zhao, Ziwei and Xu, Wujiang and Huang, Jingyuan and Guo, Minghao and Wang, Qifan and Zhou, Kaixiong and Zhang, Yongfeng},
  journal={arXiv preprint arXiv:2503.03313},
  year={2025}
}

@article{wang2025model,
  title={Model Generalization on Text Attribute Graphs: Principles with Large Language Models},
  author={Wang, Haoyu and Liu, Shikun and Wei, Rongzhe and Li, Pan},
  journal={arXiv preprint arXiv:2502.11836},
  year={2025}
}

@article{guan2025attention,
  title={Attention Mechanisms Perspective: Exploring LLM Processing of Graph-Structured Data},
  author={Guan, Zhong and Wu, Likang and Zhao, Hongke and He, Ming and Fan, Jianpin},
  journal={arXiv preprint arXiv:2505.02130},
  year={2025}
}

@inproceedings{giles1998citeseer,
  title={CiteSeer: An automatic citation indexing system},
  author={Giles, C Lee and Bollacker, Kurt D and Lawrence, Steve},
  booktitle={Proceedings of the third ACM conference on Digital libraries},
  pages={89--98},
  year={1998}
}

@article{sen2008collective,
  title={Collective classification in network data},
  author={Sen, Prithviraj and Namata, Galileo and Bilgic, Mustafa and Getoor, Lise and Galligher, Brian and Eliassi-Rad, Tina},
  journal={AI magazine},
  volume={29},
  number={3},
  pages={93--93},
  year={2008}
}

@article{craven1998learning,
  title={Learning to extract symbolic knowledge from the World Wide Web},
  author={Craven, Mark and DiPasquo, Dan and Freitag, Dayne and McCallum, Andrew and Mitchell, Tom and Nigam, Kamal and Slattery, Se{\'a}n},
  journal={AAAI/IAAI},
  volume={3},
  number={3.6},
  pages={2},
  year={1998}
}

@article{platonov2023critical,
  title={A critical look at the evaluation of GNNs under heterophily: Are we really making progress?},
  author={Platonov, Oleg and Kuznedelev, Denis and Diskin, Michael and Babenko, Artem and Prokhorenkova, Liudmila},
  journal={arXiv preprint arXiv:2302.11640},
  year={2023}
}

@misc{zheng2023judging,
      title={Judging LLM-as-a-judge with MT-Bench and Chatbot Arena},
      author={Lianmin Zheng and Wei-Lin Chiang and Ying Sheng and Siyuan Zhuang and Zhanghao Wu and Yonghao Zhuang and Zi Lin and Zhuohan Li and Dacheng Li and Eric. P Xing and Hao Zhang and Joseph E. Gonzalez and Ion Stoica},
      year={2023},
      eprint={2306.05685},
      archivePrefix={arXiv},
      primaryClass={cs.CL}
}

@inproceedings{wolf-etal-2020-transformers,
    title = "Transformers: State-of-the-Art Natural Language Processing",
    author = "Thomas Wolf and Lysandre Debut and Victor Sanh and Julien Chaumond and Clement Delangue and Anthony Moi and Pierric Cistac and Tim Rault and Rémi Louf and Morgan Funtowicz and Joe Davison and Sam Shleifer and Patrick von Platen and Clara Ma and Yacine Jernite and Julien Plu and Canwen Xu and Teven Le Scao and Sylvain Gugger and Mariama Drame and Quentin Lhoest and Alexander M. Rush",
    booktitle = "Proceedings of the 2020 Conference on Empirical Methods in Natural Language Processing: System Demonstrations",
    month = oct,
    year = "2020",
    address = "Online",
    publisher = "Association for Computational Linguistics",
    url = "https://www.aclweb.org/anthology/2020.emnlp-demos.6",
    pages = "38--45"
}

@inproceedings{Fey/Lenssen/2019,
  title={Fast Graph Representation Learning with {PyTorch Geometric}},
  author={Fey, Matthias and Lenssen, Jan E.},
  booktitle={ICLR Workshop on Representation Learning on Graphs and Manifolds},
  year={2019},
}

@article{gasteiger2019diffusion,
  title={Diffusion improves graph learning},
  author={Gasteiger, Johannes and Wei{\ss}enberger, Stefan and G{\"u}nnemann, Stephan},
  journal={Advances in neural information processing systems},
  volume={32},
  year={2019}
}

@book{Ramsundar-et-al-2019,
    title={Deep Learning for the Life Sciences},
    author={Bharath Ramsundar and Peter Eastman and Patrick Walters and Vijay Pande and Karl Leswing and Zhenqin Wu},
    publisher={O'Reilly Media},
    note={\url{https://www.amazon.com/Deep-Learning-Life-Sciences-Microscopy/dp/1492039837}},
    year={2019}
}

@article{jiao2019tinybert,
  title={Tinybert: Distilling bert for natural language understanding},
  author={Jiao, Xiaoqi and Yin, Yichun and Shang, Lifeng and Jiang, Xin and Chen, Xiao and Li, Linlin and Wang, Fang and Liu, Qun},
  journal={arXiv preprint arXiv:1909.10351},
  year={2019}
}

@article{touvron2023llama,
  title={Llama 2: Open foundation and fine-tuned chat models},
  author={Touvron, Hugo and Martin, Louis and Stone, Kevin and Albert, Peter and Almahairi, Amjad and Babaei, Yasmine and Bashlykov, Nikolay and Batra, Soumya and Bhargava, Prajjwal and Bhosale, Shruti and others},
  journal={arXiv preprint arXiv:2307.09288},
  year={2023}
}

@article{DBLP:journals/corr/abs-1907-11692,
  author    = {Yinhan Liu and
               Myle Ott and
               Naman Goyal and
               Jingfei Du and
               Mandar Joshi and
               Danqi Chen and
               Omer Levy and
               Mike Lewis and
               Luke Zettlemoyer and
               Veselin Stoyanov},
  title     = {RoBERTa: {A} Robustly Optimized {BERT} Pretraining Approach},
  journal   = {CoRR},
  volume    = {abs/1907.11692},
  year      = {2019},
  url       = {http://arxiv.org/abs/1907.11692},
  archivePrefix = {arXiv},
  eprint    = {1907.11692},
  bibsource = {dblp computer science bibliography, https://dblp.org}
}

@article{DBLP:journals/corr/abs-1810-04805,
  author    = {Jacob Devlin and
               Ming{-}Wei Chang and
               Kenton Lee and
               Kristina Toutanova},
  title     = {{BERT:} Pre-training of Deep Bidirectional Transformers for Language
               Understanding},
  journal   = {CoRR},
  volume    = {abs/1810.04805},
  year      = {2018},
  url       = {http://arxiv.org/abs/1810.04805},
  archivePrefix = {arXiv},
  eprint    = {1810.04805},
  bibsource = {dblp computer science bibliography, https://dblp.org}
}

@article{2020t5,
  author  = {Colin Raffel and Noam Shazeer and Adam Roberts and Katherine Lee and Sharan Narang and Michael Matena and Yanqi Zhou and Wei Li and Peter J. Liu},
  title   = {Exploring the Limits of Transfer Learning with a Unified Text-to-Text Transformer},
  journal = {Journal of Machine Learning Research},
  year    = {2020},
  volume  = {21},
  number  = {140},
  pages   = {1-67},
  url     = {http://jmlr.org/papers/v21/20-074.html}
}

@inproceedings{zhou-etal-2025-taming,
    title = "Taming Language Models for Text-attributed Graph Learning with Decoupled Aggregation",
    author = "Zhou, Chuang  and
      Wang, Zhu  and
      Chen, Shengyuan  and
      Du, Jiahe  and
      Zheng, Qiyuan  and
      Xu, Zhaozhuo  and
      Huang, Xiao",
    editor = "Che, Wanxiang  and
      Nabende, Joyce  and
      Shutova, Ekaterina  and
      Pilehvar, Mohammad Taher",
    booktitle = "Proceedings of the 63rd Annual Meeting of the Association for Computational Linguistics (Volume 1: Long Papers)",
    month = jul,
    year = "2025",
    address = "Vienna, Austria",
    publisher = "Association for Computational Linguistics",
    url = "https://aclanthology.org/2025.acl-long.173/",
    pages = "3463--3474",
    ISBN = "979-8-89176-251-0"
}

@inproceedings{ge-etal-2025-graph,
    title = "Can Graph Descriptive Order Affect Solving Graph Problems with {LLM}s?",
    author = "Ge, Yuyao  and
      Liu, Shenghua  and
      Bi, Baolong  and
      Wang, Yiwei  and
      Mei, Lingrui  and
      Feng, Wenjie  and
      Chen, Lizhe  and
      Cheng, Xueqi",
    editor = "Che, Wanxiang  and
      Nabende, Joyce  and
      Shutova, Ekaterina  and
      Pilehvar, Mohammad Taher",
    booktitle = "Proceedings of the 63rd Annual Meeting of the Association for Computational Linguistics (Volume 1: Long Papers)",
    month = jul,
    year = "2025",
    address = "Vienna, Austria",
    publisher = "Association for Computational Linguistics",
    url = "https://aclanthology.org/2025.acl-long.321/",
    pages = "6404--6420",
    ISBN = "979-8-89176-251-0"
}

@inproceedings{cao-etal-2025-graphinsight,
    title = "{G}raph{I}nsight: Unlocking Insights in Large Language Models for Graph Structure Understanding",
    author = "Cao, Yukun  and
      Han, Shuo  and
      Gao, Zengyi  and
      Ding, Zezhong  and
      Xie, Xike  and
      Zhou, S Kevin",
    editor = "Che, Wanxiang  and
      Nabende, Joyce  and
      Shutova, Ekaterina  and
      Pilehvar, Mohammad Taher",
    booktitle = "Proceedings of the 63rd Annual Meeting of the Association for Computational Linguistics (Volume 1: Long Papers)",
    month = jul,
    year = "2025",
    address = "Vienna, Austria",
    publisher = "Association for Computational Linguistics",
    url = "https://aclanthology.org/2025.acl-long.591/",
    pages = "12096--12134",
    ISBN = "979-8-89176-251-0"
}

@inproceedings{zhang-etal-2025-galla,
    title = "{GALL}a: Graph Aligned Large Language Models for Improved Source Code Understanding",
    author = "Zhang, Ziyin  and
      Yu, Hang  and
      Lee, Sage  and
      Di, Peng  and
      Li, Jianguo  and
      Wang, Rui",
    editor = "Che, Wanxiang  and
      Nabende, Joyce  and
      Shutova, Ekaterina  and
      Pilehvar, Mohammad Taher",
    booktitle = "Proceedings of the 63rd Annual Meeting of the Association for Computational Linguistics (Volume 1: Long Papers)",
    month = jul,
    year = "2025",
    address = "Vienna, Austria",
    publisher = "Association for Computational Linguistics",
    url = "https://aclanthology.org/2025.acl-long.676/",
    pages = "13784--13802",
    ISBN = "979-8-89176-251-0"
}

@article{ahmad2025opencodereasoning,
      title={{OpenCodeReasoning: Advancing Data Distillation for Competitive Coding}}, 
      author={Wasi Uddin Ahmad and Sean Narenthiran and Somshubra Majumdar and Aleksander Ficek and Siddhartha Jain and Jocelyn Huang and Vahid Noroozi and Boris Ginsburg},
      year={2025},
      eprint={2504.01943},
      archivePrefix={arXiv},
      primaryClass={cs.CL},
      url={https://arxiv.org/abs/2504.01943}, 
}

@article{
huang2024can,
title={Can {LLM}s Effectively Leverage Graph Structural Information through Prompts, and Why?},
author={Jin Huang and Xingjian Zhang and Qiaozhu Mei and Jiaqi Ma},
journal={Transactions on Machine Learning Research},
issn={2835-8856},
year={2024},
url={https://openreview.net/forum?id=L2jRavXRxs},
note={}
}

@inproceedings{wu2025llmnodebed,
      title={When Do LLMs Help With Node Classification? A Comprehensive Analysis}, 
      author={Xixi Wu and Yifei Shen and Fangzhou Ge and Caihua Shan and Yizhu Jiao and Xiangguo Sun and Hong Cheng},
      year={2025},
      booktitle={International Conference on Machine Learning},
      organization={PMLR},
      url={https://arxiv.org/abs/2502.00829}, 
}

@inproceedings{zhang2022protein,
  title={Protein representation learning by geometric structure pretraining},
  author={Zhang, Zuobai and Xu, Minghao and Jamasb, Arian and Chenthamarakshan, Vijil and Lozano, Aurelie and Das, Payel and Tang, Jian},
  booktitle={International Conference on Learning Representations},
  year={2023}
}

@article{10.1038/s42256-022-00580-7,
  year = {2022},
  title = {{Large-scale chemical language representations capture molecular structure and   properties}},
  author = {Ross, Jerret and Belgodere, Brian and Chenthamarakshan, Vijil and Padhi, Inkit and   Mroueh, Youssef and Das, Payel},
  journal = {Nature Machine Intelligence},
  doi = {10.1038/s42256-022-00580-7},
  pages = {1256--1264},
  number = {12},
  volume = {4}
}

@article{lin2022language,
  title={Language models of protein sequences at the scale of evolution enable accurate structure prediction},
  author={Lin, Zeming and Akin, Halil and Rao, Roshan and Hie, Brian and Zhu, Zhongkai and Lu, Wenting and Smetanin, Nikita and dos Santos Costa, Allan and Fazel-Zarandi, Maryam and Sercu, Tom and Candido, Sal and others},
  journal={bioRxiv},
  year={2022},
  publisher={Cold Spring Harbor Laboratory}
}

@Misc{Bhatia16,
          author    = {Bhatia, K. and Dahiya, K. and Jain, H. and Kar, P. and Mittal, A. and Prabhu, Y. and Varma, M.},
          title     = {The extreme classification repository: Multi-label datasets and code},
          url       = {http://manikvarma.org/downloads/XC/XMLRepository.html},
          year      = {2016}
        }

@article{wang2020microsoft,
  title={Microsoft academic graph: When experts are not enough},
  author={Wang, Kuansan and Shen, Zhihong and Huang, Chiyuan and Wu, Chieh-Han and Dong, Yuxiao and Kanakia, Anshul},
  journal={Quantitative Science Studies},
  volume={1},
  number={1},
  pages={396--413},
  year={2020},
  publisher={MIT Press One Rogers Street, Cambridge, MA 02142-1209, USA journals-info~…}
}

\appendix
\section{Dataset Details}
\label{app:dataset}
In this section, we will introduce our used datasets in details:
\begin{itemize}
    \item Cora: The Cora dataset is a classic citation network where each node represents a machine learning research paper, and edges indicate citation relationships between papers. Each paper is described by a sparse bag-of-words feature vector, and the task is to classify papers into one of seven predefined categories such as neural networks or case-based reasoning. Total 2,708 nodes will be classified into $\{$'Theory', 'Neural Networks', 'Probabilistic Methods', 'Reinforcement Learning', 'Case Based', 'Rule Learning', 'Genetic Algorithms'$\}$
    \item Citeseer: Citeseer is another widely-used citation network dataset in which nodes represent research papers and edges denote citation links. Each node includes word-based features and belongs to one of six scientific categories. These labels $\{$'artificial intelligence', 'human-computer interaction', 'information retrieval', 'database', 'agents', 'machine learning'$\}$ will be associated to 3,186 nodes in Citeseer.
    \item Pubmed: The Pubmed dataset is a large-scale citation graph composed of scientific papers from the biomedical domain. Each node represents a paper described by a TF/IDF-weighted word vector from the paper’s abstract, and edges correspond to citation links. Pubmed contains 19,717 nodes, and nodes are partitioned into 3 label categories: $\{$Diabetes Mellitus Type1, Diabetes Mellitus Type2, Diabetes Mellitus Experimental$\}$
    \item School: School dataset is a collection of 4 common heterophilic graph datasets: Cornell, Texas, Washington, and Wisconsin. All of these 4 datasets are from the WebKB collection, where represent web pages from $\{$Cornell University, University of Texas, University of Washington, University of Wisconsin$\}$ correspondingly and edges capture hyperlinks between them. Model needs to classify each node (webpage) into 5 categories: 'project', 'course', 'student', 'faculty', 'staff', and 'student'. The total number of nodes in School dataset is 872.
    \item Roman Empire: Roman Empire dataset is a synthetic temporal graph dataset designed to evaluate temporal graph learning models. There are 17 labels in total: $\{$'passive subject', 'coordinating conjunction', 'active subject', 'object of preposition', 'adverbial modifier', 'adjective modifier', 'relative clause', 'noun compound modifier', 'appositive modifier', 'prepositional marker', 'passive auxiliary', 'possessive modifier', 'direct object', 'null', 'conjoined element', 'auxiliary verb', 'main predicate', 'determiner'$\}$, and Roman Empire contains 24,492 nodes.
    \item Amazon Ratings: The Amazon Ratings dataset represents a temporal bipartite graph where nodes are users and products, and edges correspond to product ratings over time. There are 24,492 comments with 5 different rating scales: $\{$'excellent – exceeded all expectations', 'very good – almost perfect, just shy of excellent', 'decent – some good, some bad', 'good – solid experience with minor flaws', 'terrible – extremely disappointing'$\}$
    \item BACE: contains bioactivity data for small molecules that inhibit human $\beta$-secretase 1 (BACE-1), a key target in Alzheimer’s disease drug discovery. Each molecule is labeled as active or inactive, making it a binary classification task for molecular binding affinity.
    \item BBBP: consists of compounds labeled according to their ability to penetrate the blood-brain barrier, which is crucial in central nervous system drug design. This is also a binary classification task.
    \item HIV: includes information on over 40,000 compounds tested for their ability to inhibit HIV replication. Each molecule is labeled as active or inactive against HIV, making it another binary classification problem aimed at identifying potential antiretroviral candidates.
\end{itemize}
Each dataset follow the same train-test split ratio 8:2.
\section{Experiment Configuration}
\label{app:exp}
\begin{table}[!h]
    \centering
    \fontsize{9.5}{12}\selectfont
    \begin{tabular}{c|c|c}
        dataset & training epoch & total training time \\
        \midrule
        Cora & 5 & $\sim16$mins \\
        Citeseer & 5 & $\sim10$mins \\
        Pubmed & 1 & $\sim9$mins \\
        School & 13 & $\sim3$mins \\
        Roman Empire & 1 & $\sim10$mins \\
        Amazon Ratings & 1 & $\sim10$mins \\
        BACE & 12 & $\sim 5$mins \\
        BBBP & 8 & $\sim 5$mins \\
        HIV & 3 & $\sim 8$mins \\
    \end{tabular}
    \caption{Configuration and efficiency estimation for each dataset.}
    \label{tab:my_label}
\end{table}
Each dataset is trained on 8 A6000 GPUs, and the training batch size is set to 4 per GPU for all dataset, and the learning rate for template-based encoding is 2e-3 and for GNN-based encoding is 1e-4. We use AdamW optimizer and DeepSpeed to perform the multi-GPU training. We use the vicuna-7b~\citep{zheng2023judging} as our main LLM backbone for all experiments. We report average results from 3 random seed runs. For GraphToken experiments, we set the number of adapter layer at 1 for each adapter module. Setting adapter layer at 1 usually offers the best performance, and model will easily lose its expressivity with a deeper adapter layer. All models and experiments are built using Hugging Face~\citep{wolf-etal-2020-transformers} and torch geometric~\citep{Fey/Lenssen/2019} packages. 

\section{Prompts}
\label{app:prompts}
\begin{itemize}
    \item Cora: Given a node-centered graph: $<\text{graph}>$, each node represents a paper, we need to classify the center node into 7 classes: Case Based, Genetic Algorithms, Neural Networks, Probabilistic Methods, Reinforcement Learning, Rule Learning, Theory, please tell me which class the center node belongs to?
    \item Citeseer: Given a node-centered graph: $<\text{graph}>$, each node represents a paper, we need to classify the center node into 6 classes: Agents, Machine Learning, Information Retrieval, Database, Human-Computer Interaction, Artificial Intelligence, please tell me which class the center node belongs to?
    \item Pubmed: Given a node-centered graph: $<\text{graph}>$, each node represents a paper about Diabetes, we need to classify the center node into 3 classes: Diabetes Mellitus Experimental, Diabetes Mellitus Type1, Diabetes Mellitus Type2, please tell me which class the center node belongs to?
    \item School: In a graph of a university website, each node represents a web page, and each edge indicates that one web page links to another via a hyperlink. The web pages can belong to one of the following categories: project, faculty, course, student, staff. Here is a node-centered graph: $<\text{graph}>$, what is the category?
    \item Roman Empire: In an article, words that have dependency relationships (where one word depends on another) are connected, forming a dependency graph. Based on the connections between words, determine the syntactic role of each word. Given that a word described in a node-centered graph: $<\text{graph}>$,  what is this word syntactic role?
    \item Amazon Ratings: n a product graph dataset, edges connect products that are frequently purchased together. Based on the connections between products (books, music CDs, DVDs, VHS tapes), predict the average rating given by reviewers for the products. Given that a product described in a node-centered graph: $<\text{graph}>$, what is the product rating?
    \item BACE: Given the following molecule $<\text{graph}>$, determine whether it is active or inactive as a BACE-1 inhibitor.
    \item BBBP: Determine whether the following molecule $<\text{graph}>$ can penetrate the blood-brain barrier (BBB) based on its SMILES representation.
    \item HIV: This molecule is represented by the following $<\text{graph}>$. Predict whether it is active or inactive against HIV replication.
\end{itemize}
The $<\text{graph}>$ serves as a placeholder token, which will be replace by the input node sequence during training and inference stages.

\end{document}